%% file: main.tex
\documentclass[10pt,twocolumn,letterpaper]{article}

\usepackage{wacv}              


\definecolor{wacvblue}{rgb}{0.21,0.49,0.74}
\usepackage[pagebackref,breaklinks,colorlinks,allcolors=wacvblue]{hyperref}

\usepackage{adjustbox}
\usepackage{amsfonts}       
\usepackage{nicefrac}       
\usepackage{microtype}      
\usepackage{xcolor}         
\usepackage{multirow}
\usepackage{wrapfig}
\usepackage{float, bm}
\usepackage{epstopdf}
\usepackage{graphics}
\usepackage{multirow}
\usepackage{amssymb,amsmath}
\usepackage{xspace}
\usepackage[para]{footmisc}
\usepackage{enumitem}
\usepackage{placeins}
\usepackage{arydshln}
\usepackage{caption}
\usepackage{enumitem}
\usepackage{pifont}
\usepackage{tikz}
\usepackage{calc}
\usepackage{colortbl}

\newcommand{\ours}{\textsc{RONIN}\xspace}
\newcommand{\ourslong}{Ze\textbf{R}o-shot C\textbf{ON}ditional \textbf{IN}painting (\textbf{RONIN})\xspace}

\usepackage[capitalize]{cleveref}
\crefname{section}{Sec.}{Secs.}
\Crefname{section}{Section}{Sections}
\Crefname{table}{Table}{Tables}
\crefname{table}{Tab.}{Tabs.}

\def\wacvPaperID{265} 
\def\confName{WACV}
\def\confYear{2026}

\begin{document}

\title{Detecting Out-of-Distribution Objects through Class-Conditioned Inpainting}

\author{
Quang-Huy Nguyen$^{1*}$ \quad
Jin Peng Zhou$^{2*}$ \quad
Zhenzhen Liu$^{2*}$ \\
Khanh-Huyen Bui$^{3}$ \quad
Kilian Q. Weinberger$^{2}$ \quad
Wei-Lun Chao$^{1}$ \quad
Dung-Le Duy$^{4}$ \\
{\tt\small nguyen.2959@osu.edu, jz563@cornell.edu, zl535@cornell.edu} \\
{\tt\small huyenbk2@fpt.com, kilian@cornell.edu, chao.209@osu.edu, dung.ld@vinuni.edu.vn} \\
$^{1}$The Ohio State University, Columbus, Ohio, USA \quad
$^{2}$Cornell University, Ithaca, New York, USA \\
$^{3}$FPT Software AI Center, Hanoi, Vietnam \quad
$^{4}$VinUniversity, Hanoi, Vietnam \\
$^{*}$Equal contribution
}
\maketitle

\input{Sections/abstract}

\input{Sections/introduction}

\input{Sections/related_work}

\input{Sections/problem_formulation}

\input{Sections/main_method}

\input{Sections/experiment}

\input{Sections/near_OOD}

\input{Sections/ablation}

\input{Sections/conclusion}

\newpage

{\small
\bibliographystyle{ieee_fullname}
\bibliography{reference}
}

\newpage
\appendix
\onecolumn

\input{Sections/appendix}

\end{document}

%% file: Sections/abstract.tex
\begin{abstract}
Recent object detectors have achieved impressive accuracy in identifying objects seen during training. However, real-world deployment often introduces novel and unexpected objects, referred to as out-of-distribution (OOD) objects, posing significant challenges to model trustworthiness. Modern object detectors are typically overconfident, making it unreliable to use their predictions \emph{alone} for OOD detection. To address this, we propose leveraging an \emph{auxiliary} model as a complementary solution. Specifically, we utilize an off-the-shelf text-to-image generative model, such as Stable Diffusion, which is trained with objective functions distinct from those of discriminative object detectors. We hypothesize that this fundamental difference enables the detection of OOD objects by measuring inconsistencies between the models. Concretely, for a given detected object bounding box and its predicted in-distribution class label, we perform class-conditioned inpainting on the image with the object removed. If the object is OOD, the inpainted image is likely to deviate significantly from the original, making the reconstruction error a robust indicator of OOD status. Extensive experiments demonstrate that our approach consistently surpasses existing zero-shot and non-zero-shot OOD detection methods, establishing a robust framework for enhancing object detection systems in dynamic environments. Our implementation is available at \url{https://github.com/quanghuy0497/RONIN}.
\end{abstract}

%% file: Sections/introduction.tex
\section{Introduction}
\label{sec:introduction}

\begin{figure}[ht!]
  \centering
  \includegraphics[width=.48\textwidth]{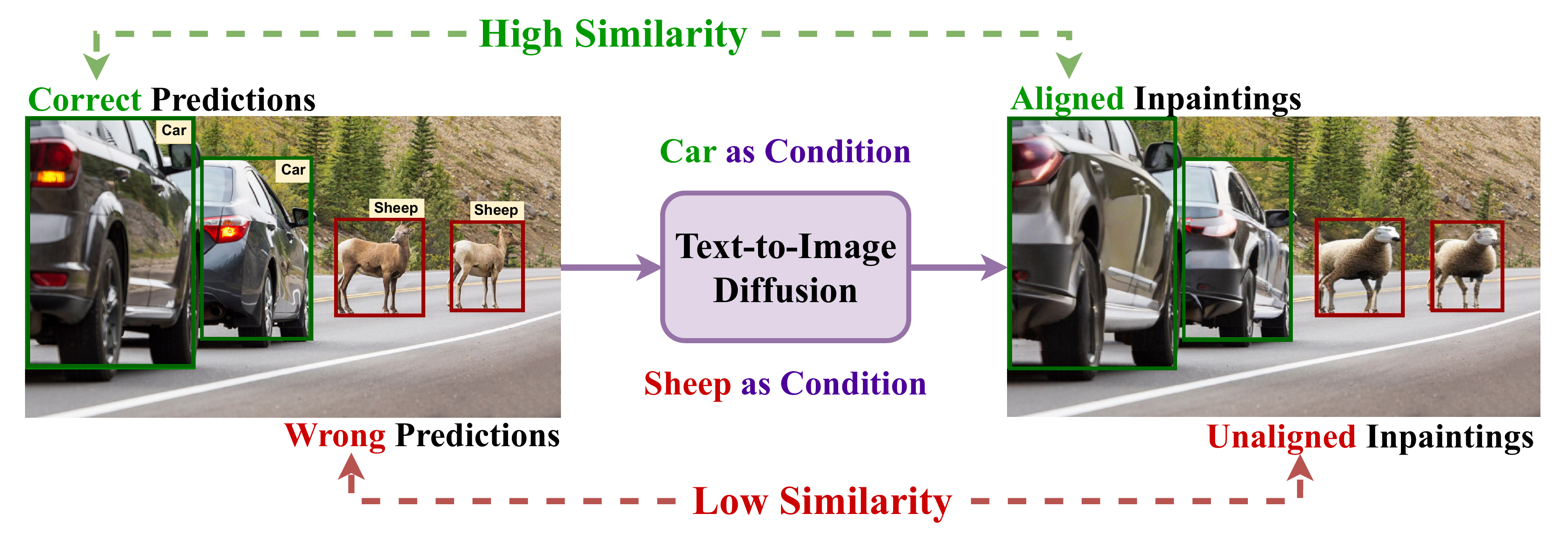}
\caption{\textbf{Intuition behind \ours for OOD detection}. Object detectors can overconfidently make wrong predictions on unseen objects based on a pre-defined set of ID labels; in this case, predicting two OOD \textit{deers} as ``\textit{sheep}''. \ours leverages the label predictions to condition the resynthesizing process of an off-the-shelf text-to-image diffusion model, producing similar inpaintings for correct predictions and dissimilar inpaintings for incorrect ones. Under a similarity measurement, \ours can identify the wrong prediction of unseen objects, therefore ODD detection.}
    \label{fig:intuition}
\vspace{-1.5em}
\end{figure}

Object detection systems have achieved remarkable progress in recent years, becoming integral to a wide range of applications in both online and offline settings. These include, but are not limited to, environmental monitoring~\cite{beery2019efficient}, industrial manufacturing~\cite{ahmad2022deep}, and healthcare~\cite{Ragab2024ACS}. As these systems increasingly influence decision-making processes, ensuring their reliability and robustness has become a key challenge. 

One common failure mode arises when object detectors encounter object categories not seen during training, referred to as out-of-distribution (OOD) cases. Such occurrences are inevitable due to the dynamic and ever-changing nature of the real world. Unfortunately, modern object detectors often exhibit overconfidence, frequently misclassifying OOD objects as belonging to in-distribution (ID) categories, thereby resulting in false positives~\cite{pathiraja2023multicalib}. This overconfidence underscores a fundamental limitation: the internal responses of object detectors cannot reliably identify these errors. Consequently, alternative strategies are needed to enhance the trustworthiness of these systems.

A popular strategy is to ``rehearse'' these error cases during detector training. For example, \cite{munir2022towards, pathiraja2023multicalib} introduced auxiliary losses to regularize model confidence, while \cite{du2022vos, du2022siren} proposed methods to learn more distinguishable detector features between ID and OOD categories. While promising, these strategies require retraining and are not widely implemented in pre-trained, publicly accessible object detectors. Given that many stakeholders rely on off-the-shelf detectors for their applications, there is a strong demand for effective \emph{post-hoc} solutions to address OOD detection challenges.

In this paper, we propose leveraging an additional pre-trained model to provide auxiliary information for effective OOD detection. Specifically, we utilize off-the-shelf text-conditioned generative models~\cite{rombach2021highresolution,saharia2022imagen}, which are trained to synthesize realistic images from text prompts, such as category names. We hypothesize that the distinct training objectives of object detectors and generative models lead to inconsistent responses for OOD cases. These inconsistencies thus serve as a robust post-hoc indicator for OOD detection, offering a practical and accessible solution to enhance the reliability of object detection systems without requiring retraining.

More specifically, given a detected object bounding box and its predicted ID class label by the object detector, we propose performing class-conditioned inpainting for OOD detection. We mask the image portion within the box and use the ID class label as the prompt to resynthesize it. If the masked-out object is ID, the reconstructed image portion should be similar to the original, both visually and semantically. In contrast, for an OOD object, the inpainted portion prompted by the misclassified ID label will significantly deviate from the original. This discrepancy provides a valuable cue for distinguishing ID and OOD objects, requiring neither retraining nor access to ID or OOD training examples. We name our approach \ourslong, as illustrated in \cref{fig:intuition}.

We evaluate \ours on real-world benchmark datasets, including PASCAL VOC~\cite{everingham2010pascal}, BDD-100K~\cite{yu2020bdd100k}, MS-COCO~\cite{lin2014microsoft}, and OpenImages~\cite{kuznetsova2020open}, covering indoor, outdoor, and in-the-wild scenarios for comprehensive evaluation. The empirical results demonstrate \ours's superior OOD detection performance compared to existing methods. Notably, \ours shows strong potential in near-OOD scenarios by incorporating potential nearest non-ID concepts as exclusions, which leads to larger reconstruction errors when the original object belongs to one of these concepts and thus address near-OOD detection. Extensive ablation studies, analyses, and visualizations further validate the effectiveness of \ours.

In summary, \ours offers three notable advantages, positioning itself as a \emph{plug-and-play} solution for OOD detection:
\begin{itemize}[nosep, topsep=2pt, parsep=2pt, partopsep=2pt, leftmargin=*]
\item \textbf{Retraining-free and zero-shot:} \ours requires no access to training data for either ID or OOD categories, making it \textit{highly versatile} and \textit{data-efficient}.
\item \textbf{Ease of implementation and compatibility:} \ours is \textit{straightforward to implement} and \textit{compatible with most existing object detectors}, including open-vocabulary object detectors~\cite{liu2023grounding}, which still rely on a pre-defined set of ID concepts to limit their scope.
\item \textbf{Leveraging off-the-shelf generative models:} By utilizing pre-trained generative models, \ours continuously benefits from advancements in generative modeling, \textit{ensuring improved effectiveness and efficiency over time}.
\end{itemize}

\noindent\textbf{Remark.}~~~While straightforward to implement, \ours is far from a trivial contribution. It effectively leverages the inconsistency between discriminative and generative models as a post-hoc signal for identifying OOD objects. Unlike~\cite{liu2023lmd}, \ours extends applicability to object detection without requiring diffusion models to be trained on ID data. With the growing availability of powerful pre-trained models, using them as auxiliary tools to enhance OOD detection in domain-specific pipelines is both practical and scalable.

A potential limitation of \ours lies in runtime, as diffusion models are generally slower than object detectors. However, this can be mitigated by reducing denoising steps—since our focus is on detecting OOD objects, not generating high-fidelity images, a slight quality trade-off is acceptable. We investigate this in~\cref{sec:experiment}, showing that fewer denoising steps significantly reduce runtime with minimal performance drop.

Importantly, object detectors are not always used in real-time. Many applications, such as wildlife monitoring with camera traps, involve offline post-processing where accuracy and generalizability matter more than speed. Our experiments across indoor and outdoor scenes validate \ours’s strong performance in such settings, highlighting its readiness for practical deployment.

As generative modeling continues to evolve—with efficiency gains from methods like one-step diffusion—\ours stands to benefit further. By prioritizing flexibility, generalizability, and practicality in both design and empirical results, \ours offers a meaningful contribution to zero-shot out-of-distribution object detection.

%% file: Sections/related_work.tex
\section{Related Work}
\label{sec:related_work}

\noindent\textbf{Out-of-distribution Detection.}~~~The goal of the out-of-distribution (OOD) detection task is to determine whether a given sample belongs to a certain distribution. It has been widely studied at the \textit{image level}, where a whole image is treated as a sample. Common approaches include leveraging classifier-specific information such as confidence scores~\cite{hendrycks2016baseline, lee2017training, liang2018odin, devries2018learning, hsu2020generalizedodin, liu2020energy, wei2022logitnorm} or learned features~\cite{lee2018simple,denouden2018improving,tack2020csi, sehwag2021ssd, xiao2021we,Sun2022knn}, or directly modeling an image distribution using generative models~\cite{ren2019likelihood, serra2019input, xiao2020likelihoodregret, schlegl2017unsupervised,zong2018deep, Graham2023denoising, liu2023lmd, li2023rethinking}. Recent works~\cite{ming2022delving, esmaeilpour2022zero, wang2023clipn} have also explored using CLIP~\cite{radford2021clip} to identify OOD examples in a zero-shot manner, bypassing the need to learn from in-domain data explicitly. 

OOD detection can also be extended to the \textit{object-level}, where objects within an image are treated as individual samples. Most existing research in this setting focuses on training-time interventions. For instance, \cite{du2022vos,du2022siren} improve detector features to make them more separable between in-distribution (ID) and OOD data; \cite{wilson2023safe} uses adversarial examples to train an MLP for classifying ID and OOD instances. In contrast, our approach \ours explores the object-level setting through post-hoc interventions instead. 



\noindent\textbf{Text-to-Image Generative Models.}~~~Recent advances in generative modeling have made large-scale text-to-image models widely available~\cite{rombach2021highresolution,ramesh2022dalle,saharia2022imagen}. These models exhibit a deep understanding of language and are highly effective at generating or editing high-quality images from diverse prompts, offering a promising approach to data synthesis across various tasks. Furthermore, recent works have shown that these models can also enhance discriminative tasks. For instance, \cite{li2023your, jaini2023intriguing} demonstrate that Stable Diffusion~\cite{rombach2021highresolution} can function as an effective zero-shot classifier, achieving accuracy comparable to or surpassing CLIP and various trained discriminative classifiers. There has also been recent progress in applying text-conditioned diffusion models for image-level OOD detection~\cite{du2023dream, gao2023diffguard, fang2024ddos}. Unlike these works, \ours focuses on exploring the use of such models for object-level tasks. 


%% file: Sections/problem_formulation.tex
\section{Problem Formulation}
\label{sec:problem_formulation}

We address the task of object-level out-of-distribution (OOD) detection, which is motivated by the fact that object detectors can recognize unseen objects as seen ones. Specifically, given an object detector trained to detect a predefined set of categories (\eg different kinds of vehicles), we aim to identify the error cases when it wrongly recognizes a novel object (\eg a wild animal) as one of the predefined categories and detects it overconfidently. We refer to the pre-defined set of categories as the in-distribution (ID) classes, following~\cite{du2022vos}; the novel categories as the OOD classes. 

Formally, given an image $\bm x$ and an object detector $f(.;\ \theta)$ trained for the ID classes $\mathcal{Y}^{in}$, $f(\bm x;\ \theta)$ outputs a list of bounding boxes $\bm{b} = \{b_1,\ b_2,\dots,\ b_n\}$ and their associated ID class labels $ \bm {\hat y} = \{\hat y_1$, $\hat y_2$,$\dots$, $\hat y_n$\}, where $\hat y_i \in \mathcal{Y}^{in}$. Object-level OOD detection is then formulated as a binary classification problem, classifying whether the object within $b_i$ truly belongs to ID classes. Typically, one would develop a scoring function $\bm g$ such that given a bounding box and its predicted label $(b, \ \hat y)$, $\bm g$ gives a higher score $\bm g(b,\ \hat y)$ if $b$ outlines an ID object and a lower score if it outlines an OOD object. 

Existing works~\cite{du2022vos, du2022siren} proposed a specific training process for the object detector 
so that its responses (\eg the feature vector of each bounding box) would better distinguish between ID and OOD classes. However, doing so requires re-training the object detector, 
assuming access to the original training data and making it infeasible to off-the-shelf object detectors.

\noindent\textbf{Aim.}~~~To address this limitation, we aim to design an OOD detection mechanism that is \textbf{\textit{post-hoc}}, without the need to modify (\eg fine-tune) the pre-trained object detector, and \textbf{\textit{zero-shot}}~\cite{ming2022delving}, without the need to access the original training data or any ID-class data. This sharply contrasts several prior post-hoc methods that need ID-class data~\cite{ren2019likelihood,xiao2020likelihoodregret}. 

\noindent\textbf{Approach.}~~~The emergence of vision-language foundation models~\cite{radford2021clip,li2022blip,liu2024visual} trained on abundant images and free-form text pairs has gradually removed the boundary between the closed-set and open-set settings~\cite{li2023your}. For example, while not perfect, CLIP~\cite{radford2021clip} can match an image with unbounded concepts. Such a zero-shot capability has been leveraged in prior work to detect OOD samples given a set of ID concepts~\cite{ming2022delving, esmaeilpour2022zero, wang2023clipn}. In this work, we leverage another kind of foundation model, text-to-image generative models~\cite{rombach2021highresolution}, capable of generating images given free-form text. While not designed for object detection nor specifically optimized for the ID data, we surmise their built-in, generic capability would facilitate object-level OOD detection. 

%% file: Sections/main_method.tex
\section{RONIN: Zero-Shot OOD Conditional Inpainting}
\label{sec:method}

\begin{figure*}[ht!]
  \centering
  \includegraphics[width=\linewidth]{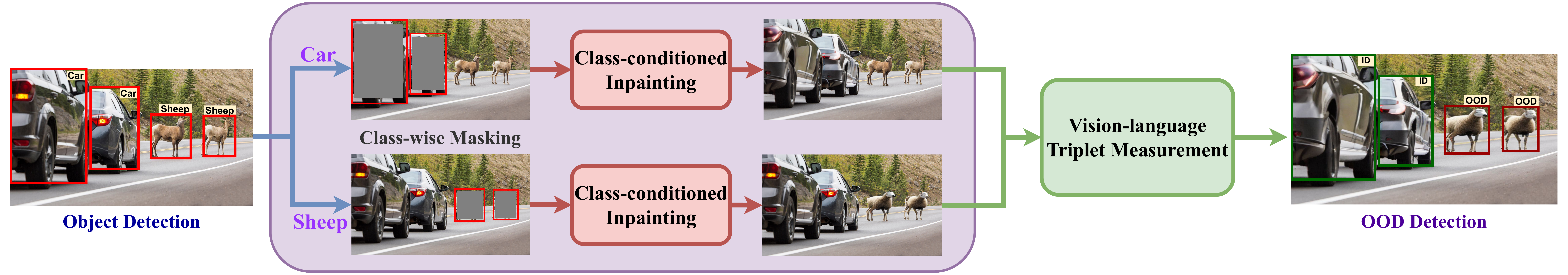}
    \caption{\textbf{Overall framework of \ours}, including (i) \textit{Class-conditioned Inpainting} and (ii) \textit{Vision-language Triplet Measurement}. Given an image with bounding boxes and predicted labels, RONIN masks and reconstructs objects via inpainting, then evaluates alignment using triplet similarity for zero-shot OOD detection.}
    \label{fig:main_method}
    \vspace{-0.2in}
\end{figure*}

Our proposed framework, \ours, builds on the assumption that generative models and object detectors behave differently when encountering unseen OOD objects due to their distinct training objectives. Object detectors are trained to separate foreground from background and classify objects into pre-defined ID categories, learning $P(y|x)$. This results in soft decision boundaries that can overconfidently assign unseen objects to seen classes. In contrast, generative models aim to capture the full data distribution $P(x, y)$ by modeling nuanced category characteristics, making them sensitive to deviations from training data and thus effective at flagging outliers. Leveraging this, we use generative models as auxiliary signals to improve zero-shot OOD detection in conjunction with object detection outputs.

\ours is illustrated in \cref{fig:main_method}. Given an object with its predicted bounding box and ID label, \ours uses off-the-shelf class-conditioned inpainting models to regenerate the object, aligning it with the in-distribution domain. The regenerated image is then compared with the original using a vision-language triplet similarity metric. High similarity implies visual and semantic consistency with the ID category (in-distribution), while low similarity suggests a mismatch (out-of-distribution). We detail each component of this approach below.

\subsection{Class-conditional Inpainting}
\label{sec:context_inpainting}

Our key idea behind \ours is to harness diffusion models for inpainting, refine predicted objects, and bring them closer to the ID domain. Specifically, given a detected bounding box $\bm{b}$ and a predicted ID class $\hat{y} \in \mathcal{Y}^{in}$ by the object detector, we generate a masked region $\bm{m}$ within $\bm{b}$ and use $\hat{y}$ as a condition for reconstruction. Since $\bm{m}$ is always contained within $\bm{b}$, the inpainting model receives sufficient context from both the background and the detected object as hints for reconstructing the original object. This ensures that the inpainted outcomes closely align with the original ID objects, not only in semantic meaning but also in visual appearance, while still being distinct in OOD cases.

More specifically, let the semantic background $\bm s$ be the entire image region excluding the masked region $\bm{m}$, \ours takes input of $\bm m$ (also known as the inpainting mask), the predicted label $\bm c = \hat y$, and the semantic background $\bm s$ for inpainting. We choose to use pre-trained text-conditioned diffusion models such as DDPM \cite{ho2020denoising}. The diffusion models
starts with a random Gaussian noise over $\bm m$ and perform the reverse diffusion process with $T$ steps, guided by $\bm c$, to obtain the outcome image $\bm{x}_{inp} = {\bm{x}}_0$, as in \cref{equa:revsere_inp}, where $\mu_\theta$ and $\Sigma_\theta$ are priorly learned by pre-training a neural network with parameter $\theta$.
\begin{align}
    p_\theta(\tilde{\bm x}_{t-1} | \bm x_t,\ \bm c) &= \mathcal{N}\big(\tilde{\bm x}_{t-1} ;\  \mu_\theta(\bm x_t, \ t, \ \bm c), \Sigma_\theta(\bm x_t, \ t, \ \bm c)\big)\label{equa:revsere_inp}\\
    \bm x_{t-1} & = \bm {\tilde x}_{t-1} \odot \bm m + \bm s
    \label{equa:masking}
\end{align}

While the inpainted objects are synthesized based on $\hat y$ over the foreground region $\bm m$, the information retained from $\bm s$ guides the DDPM to preserve some visual features of the original object, as described in \cref{equa:masking}. As a result, the inpainted outcome object retains the same visual and semantic characteristics as the original object, despite being presented differently. This enables \ours the generation of objects closely aligned with the originals in ID cases and significantly deviating in OOD cases, as shown in \cref{fig:main_method}. We further analyze how varying the inpainted mask $\bm m$ based on $\bm b$ impacts OOD detection in \cref{sec:Ablation}.

\noindent\textbf{Class-wise Masking and Inpainting.}~~~Intuitively, a natural approach to inpainting is treating each object independently, or ``\textit{object-wise}'' inpainting. However, this is highly inefficient when many objects share the same predicted label and denoising condition, making the process redundant. To address this, we propose ``\textit{class-wise}'' inpainting, where objects are grouped by the predicted label $\hat{y}$ for a single inpainting process, ensuring the computational cost at most scales with $\mathcal{Y}^{in}$ the worst case. This significantly reduces computational cost while maintaining performance, as the shared denoising condition ensures effective synthesis. We further evaluate the two strategies in \cref{sec:Ablation}.

\subsection{Vision-language Triplet Measurement}
\label{sec:similarity_measurement}

\begin{figure}[ht]
    \centering
    \includegraphics[width=\linewidth]{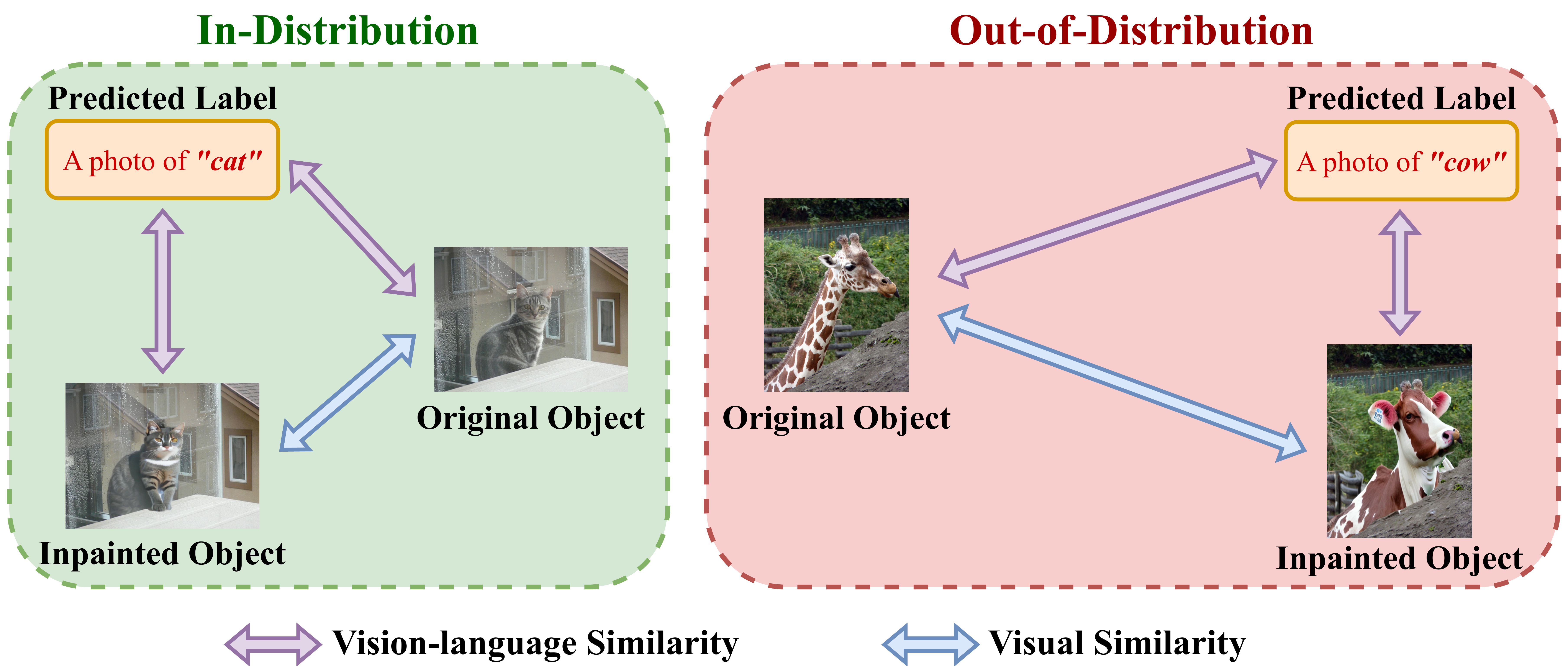}
    \vspace{-0.1in}
    \caption{\textbf{Triplet similarity relationships} between (i) the original object, (ii) the inpainted outcome, and (iii) the predicted label. ID samples show strong alignments across all three, whereas OOD samples exhibit weak alignments, aiding effective OOD detection.}   
    \label{fig:similarity_intuition}
    \vspace{-0.1in}
\end{figure}

\noindent\textbf{Similarity Aligments.}~~~After inpainting, we obtain the original object $\mathbf{x_{ori}}$, its predicted label $\hat{y} \in \mathcal{Y}^{in}$, and the corresponding inpainted object $\mathbf{x_{inp}}$ which is synthesized based on $\hat y$. Intuitively, for in-distribution (ID) objects, $\mathbf{x_{inp}}$ tends to closely resemble $\mathbf{x_{ori}}$, while in contrast, out-of-distribution (OOD) objects exhibit lower similarity, indicating divergence from the original given a wrong label.

However, due to the stochastic nature of diffusion inpainting and subtle distinctions in semantic meanings between OOD samples and predicted ID labels, the similarity between $\mathbf{x_{ori}}$ and $\mathbf{x_{inp}}$ can vary greatly. To mitigate this, we introduce a triplet of similarity scores for a more robust ID vs. OOD distinction:
\begin{equation}    
\begin{aligned}
    \textit{similarity}{(ori,\ \hat{y})} & = \mathbf{cosine}\left(f_{v}(\mathbf{x_{ori}}), f_{t}(\hat{y})\right) \\
    \textit{similarity}{(inp,\ \hat{y})} & = \mathbf{cosine}\left(f_{v}(\mathbf{x_{inp}}), f_{t}(\hat{y})\right)\\
    \textit{similarity}{(ori, \ inp)} & = \mathbf{cosine}\left(f_{i}(\mathbf{x_{ori}}), f_{i}(\mathbf{x_{inp}})\right) \
\end{aligned}
\label{equa:similarities}
\end{equation}
Here, $\mathbf{cosine}$ means the cosine similarity between two vectors and $f_v$ and $f_t$ represent visual and textual embeddings from large vision-language contrastive models (e.g., CLIP~\cite{radford2021clip}), while $f_i$ is derived from a visual contrastive model (e.g., SimCLRv2~\cite{chen2020simclrv2}). Cosine similarity effectively measures the alignment between these representations. Specifically, $\textit{similarity}{(ori, \ inp)}$ emphasizes \textit{\textbf{visual alignment}} for ID retention while $\textit{similarity}{(ori, \ \hat{y})}$ supports \textit{\textbf{semantic alignment}} for OOD recognition. In contrast, $\textit{similarity}{(inp, \ \hat{y})}$ helps \textit{\textbf{normalize}} variations across different labels. \cref{fig:similarity_intuition} illustrates the relationships between these components for both ID and OOD cases.

\noindent\textbf{Triplet Measurement.}~~~Based on
the relationship between the three similarities, we then formulate a triplet OOD score:
\begin{equation}
     \text{S}_\text{triplet} = \frac{\textit{similarity}{(ori, \ \hat{y})}^{\bm\alpha} \times \textit{similarity}{(ori, \ inp)}^{\bm\beta}}{\textit{similarity}{(inp, \ \hat{y})}}
     \label{equa:triplet_score}
\end{equation}
where $\bm\alpha$ and $\bm\beta$ are hyperparameters controlling the balance between visual-language and visual similarities. A higher $\text{S}_\text{triplet}$ suggests a greater likelihood of the object being ID, while lower values indicate OOD cases. In practice, a thresholding method can be applied for effective OOD detection depending on different tasks.

\subsection{Refined Prompt for Near-OOD}
\label{sec:method_nearOOD}
Identifying near-OOD cases poses a significant challenge for conventional OOD detection methods, as unseen objects often share closely related semantic meanings with certain in-distribution (ID) categories, making separation difficult. However, off-the-shelf generative and CLIP models inherently possess the ability to distinguish such cases because of their extensive pre-training on large, diverse datasets. To validate this, we proposed a simple approach that, given a predefined set of ID labels, we use knowledge graphs or taxonomies from large language models to retrieve closely related non-ID concepts and explicitly instruct the generative model not to inpaint these concepts as part of the conditioning. For example, if the ID label is ``\textit{horse}'', we can retrieve highly related but OOD classes such as—\textit{donkey, zebra, mule, pony, and camel}—and explicitly condition the inpainting model to generate a horse that remains distinct from these related categories. We believe this ``\textit{refined}'' prompting technique not only supports \ours to generate outcomes that align better with the ID label but also improve its ability to do near-OOD detection.

%% file: Sections/experiment.tex
\section{Experiments}
\label{sec:experiment}

\input{Sections/main_table_DETR}

\begin{figure*}[ht!]
  \centering
  \includegraphics[width=\linewidth]{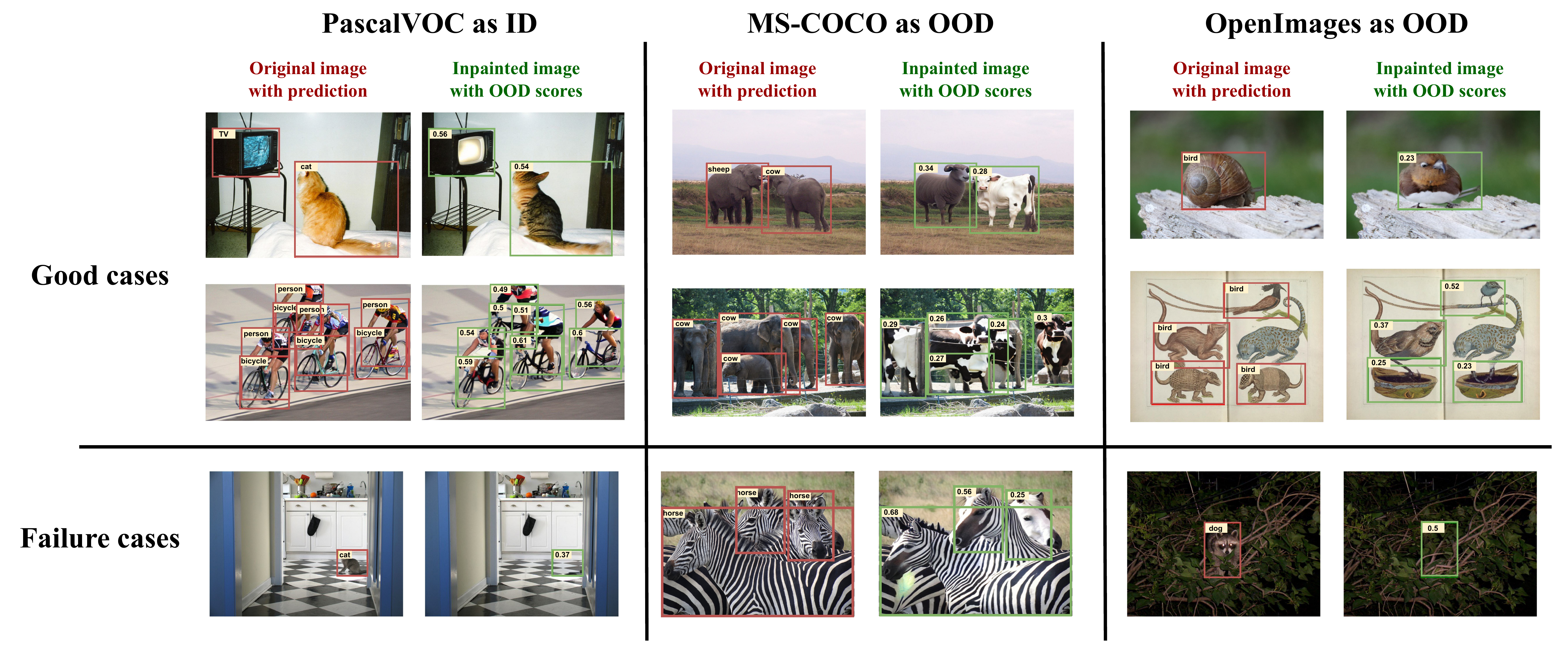}
    \caption{\textbf{Side-by-side quantitative visualization.} Good cases show synthesized objects consistent with predicted labels and clear OOD score separation; bad cases show inpainting failures or OOD objects too resembling the originals, leading to ineffective OOD score.}
    \label{fig:quantitative_performance}
    \vspace{-0.2in}
\end{figure*}

\subsection{Experimental Setups}

\noindent\textbf{Datasets.}~~~We use Pascal-VOC~\cite{everingham2010pascal} and Berkerly DeepDrive (BDD-100k)~\cite{yu2020bdd100k} as ID data, and MS-COCO~\cite{lin2014microsoft} and OpenImages~\cite{kuznetsova2020open} as OOD data. The combinations of these data covered indoor, outdoor, and in-the-wild objects and contexts, allowing us to evaluate \ours comprehensively. We follow the splits from \cite{du2022siren, du2022vos}, where \textit{overlapping categories in the ID and OOD datasets are removed}. We select a subset of 200 images from the test set of BDD-100k and 400 each from the test sets of Pascal-VOC, MS-COCO, and OpenImages. Data processing details can be found in  \cref{app:data}.

\noindent\textbf{Evaluation Metrics.}~~~We evaluate the OOD detection performance following the two standard OOD metrics: the area under the Receiver Operating Characteristic curve (\textbf{AUROC}) and the false positive rate of OOD objects at 95\% true positive rate of in-distribution objects (\textbf{FPR@95}). 

\noindent\textbf{Baselines.}~~~We consider three types of baselines: (1) \textit{Discriminative zero-shot  approaches}:  \textbf{MCM}~\cite{ming2022delving}, CLIP-based \textbf{ODIN}~\cite{liang2018odin}, CLIP-based \textbf{Energy Score}~\cite{liu2020energy}, \textbf{CLIPN}~\cite{wang2023clipn}, \textbf{TAG}~\cite{liu2024tag}, \textbf{OLE}~\cite{ding2024zero} and \textbf{GL-MCM}~\cite{miyai2025gl}. (2) \textit{Alternative generative zero-shot approaches}: We first generate synthetic data and then apply standard OOD detection methods such as \textbf{Mahalanobis}~\cite{xiao2021we} and \textbf{KNN}~\cite{Sun2022knn}. (3) \textit{Training-based approaches}: for a complete comparison, we also include two representative training-based object-level methods \textbf{VOS}~\cite{du2022vos} and \textbf{SIREN}~\cite{du2022siren}, resulting in a total of \textbf{eleven} baselines. Implementation details can be found in \cref{app:baseline_detail}

\noindent\textbf{Implementation Details.}~~~ For \ours, we perform class-wise inpainting with Stable Diffusion 2 Inpainting~\cite{rombach2021highresolution} with 20 steps. Each object is masked with a center mask covering 0.9 of the height and width of its original bounding box.  Object-label similarities are determined by the distance between OpenCLIP~\cite{gabriel2021openclip} features, and object-object similarities by the SimCLRv2~\cite{chen2020simclrv2} features. 

For the main experiments in \cref{sec:main-exp}, we use detected objects obtained from Deformable-DETR~\cite{zhu2020deformable}, as implemented by \cite{du2022siren}, to enable direct comparisons with training-based baselines. However, it is important to note that \ours is compatible with various types of detectors. In \cref{sec:nearOOD}, we demonstrate \ours’s performance using detections from the pre-trained open-vocabulary detector GroundingDINO~\cite{liu2023grounding}, showcasing its flexibility and compatibility. The Triplet Similarity (\cref{equa:triplet_score}) is computed with $\bm\alpha=2$ and $\bm\beta=1$. All experiments were conducted on a single NVIDIA GeForce RTX 2080Ti 11GB.

\subsection{Quantitative and Qualitative performance}\label{sec:main-exp}

\cref{tab:main_table_DETR} shows the performance of \ours and baselines based on DeformableDETR detections across four settings. \ours achieves the highest results in three scenarios, even with just 5 denoising steps, outperforming training-based methods like VOS and SIREN without retraining on in-domain data. Increasing the number of denoising steps further improves performance, demonstrating that \ours consistently surpasses all baselines across most settings.

\cref{fig:quantitative_performance} presents side-by-side visualizations of \ours with Pascal-VOC as the ID setting. Overall, ID objects are well reconstructed while OOD objects appear distinctly different, enabling effective separation via the triplet score. However, we identify two common types of failures: \textit{inpainting failures}, where small, occluded, or poorly lit objects prevent meaningful reconstruction, and \textit{similarity-score failures}, where generated objects remain visually or semantically too close to the original, reducing the effectiveness of the triplet score for OOD detection.

%% file: Sections/main_table_DETR.tex
\begin{table*}[ht!]
  \centering
  \caption{\textbf{Object-level OOD detection on four settings}. We highlight the \textbf{best} and \underline{second  best} performance. Even with 5 denoising steps, \ours is able to outperform all the baselines in three out of four settings. Denoising with more steps further improve the performance.}
  \begin{adjustbox}{max width=\textwidth}
  \begin{tabular}{llcccc}
    \toprule
    \multirow{2}{*}{\textbf{In-Distribution}} & \multirow{2}{*}{\textbf{Method}} 
    & \multicolumn{2}{c}{\textbf{MS-COCO}} & \multicolumn{2}{c}{\textbf{OpenImages}} \\
    \cmidrule(lr){3-4}
    \cmidrule(lr){5-6}
    & & FPR@95 $(\downarrow)$ & AUROC $(\uparrow)$ & FPR@95 $(\downarrow)$ & AUROC $(\uparrow)$ \\
    \midrule   
    \multirow{15}{*}{\textbf{PASCAL-VOC}} 
            & ODIN \cite{liang2018odin}         
            & 41.65 & 88.22 & 55.87 & 86.46 \\
            & Energy Score \cite{liu2020energy} 
            & 29.48 & 90.26 & 24.57 & 91.24 \\
            & Mahalanobis \cite{xiao2021we}     
            & 63.30 & 83.26 & 45.22 & 87.11 \\
            & KNN \cite{Sun2022knn}             
            & 58.56 & 85.52 & 45.00 & 84.62 \\
            & VOS \cite{du2022vos}              
            & 48.15 & 88.75 & 54.63 & 83.65 \\
            & SIREN \cite{du2022siren}          
            & 64.70 & 78.68 & 66.69 & 75.12 \\ 
            & MCM \cite{ming2022delving}        
            & 62.47 & 83.15 & 71.52 & 81.45 \\
            & CLIPN \cite{wang2023clipn}        
            & 43.09 & 85.45 & 41.74 & 89.31\\
            & TAG \cite{liu2024tag}             
            & 61.03 & 78.35 & 48.48 & 87.69\\
            & OLE \cite{ding2024zero}           
            & 54.23& 86.13 & 49.13 & 88.07\\
            & GL-MCM \cite{miyai2025gl}         
            & 78.56 & 78.64 & 69.35 & 82.89\\
            \cmidrule(lr){2-6}
            & \textbf{\ours} \ \ (5 steps)          
            & 27.94 & 92.35 & 20.00 & 92.32 \\
            & \textbf{\ours} (10 steps)             
            & 29.07 & 91.10 & 20.20 & 92.80 \\
            & \textbf{\ours} (15 steps)             
            & \underline{25.57} & \underline{91.63} & \underline{19.87} & \underline{92.84} \\
            & \textbf{\ours} (20 steps)             
            & \textbf{25.36} & \textbf{92.31} & \textbf{18.91} & \textbf{93.10} \\

    \midrule
    \multirow{15}{*}{\textbf{BDD-100k}} 
            & ODIN \cite{liang2018odin}         
            & 96.51 & 55.18 & 95.56 & 57.87 \\
            & Energy Score \cite{liu2020energy} 
            & 71.75 & 74.49 & 53.33 & 73.09 \\
            & Mahalanobis \cite{xiao2021we}     
            & 31.75 & 90.74 & \textbf{23.33} & \textbf{93.81} \\
            & KNN \cite{Sun2022knn}             
            & 35.87 & 86.58 & 31.11 & 92.75 \\
            & VOS \cite{du2022vos}              
            & 65.45 & 78.34 & 59.23 & 80.42 \\
            & SIREN \cite{du2022siren}          
            & 42.86 & 89.37 &  37.97 & 91.78 \\ 
            & MCM \cite{ming2022delving}        
            & 95.56 & 55.82 & 92.22 & 57.05 \\
            & CLIPN \cite{wang2023clipn}        
            & 28.49 & 92.14 & 44.76 & 85.78\\
            & TAG \cite{liu2024tag}             
            & 46.43 & 90.85 & 55.30 & 76.67 \\
            & OLE \cite{ding2024zero}           
            & 57.14 & 78.14 & 40.00 & 85.36 \\
            & GL-MCM \cite{miyai2025gl}         
            & 96.83 & 50.82 & 94.44 & 58.88\\
            \cmidrule(lr){2-6}
            & \textbf{\ours} \ \ (5 steps)          
            & 27.94 & 92.59 & 26.67 & 91.23 \\
            & \textbf{\ours} (10 steps)             
            & 26.67 & 92.73 & 25.56 & 91.33 \\
            & \textbf{\ours} (15 steps)             
            & \underline{26.03} & \underline{92.83} & 24.44 & 91.65 \\
            & \textbf{\ours} (20 steps)             
            & \textbf{26.03} & \textbf{92.90} & \textbf{23.33} & \underline{91.90} \\
    \bottomrule
  \end{tabular}
  \end{adjustbox}
  \label{tab:main_table_DETR}
\end{table*}

%% file: Sections/near_OOD.tex
\subsection{Near-OOD Detection}
\label{sec:nearOOD}

\begin{figure}[htbp]
\centering
\includegraphics[width =.95\linewidth]{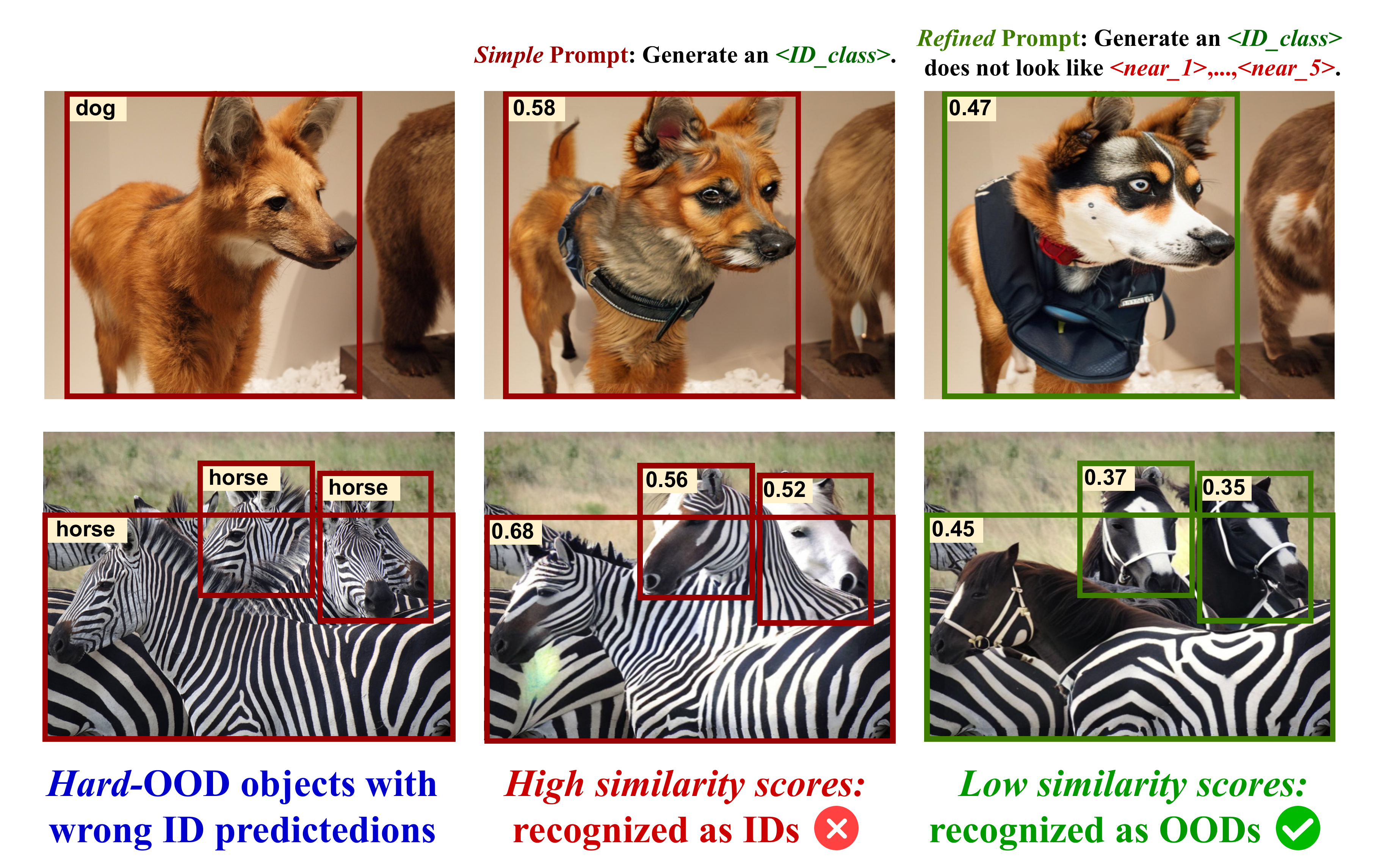}
\caption{\textbf{\ours performance on near-OOD with \textit{refined} prompting}. With distinct inpainting by providing more context, \ours is able to yield lower scores for near-OOD detection.} 
\label{fig:near_ood}  
\vspace{-2em}
\end{figure}

To demonstrate the potential of \ours for near-OOD detection, we constructed a small-scale dataset featuring visually and semantically similar ID and OOD objects. Specifically, we curated 411 samples by pairing ID objects from PascalVOC with closely related OOD counterparts from MS-COCO and OpenImages, forming categories such as ``\textit{horse vs. zebra}'', ``\textit{dog vs. fox}'', and ``\textit{cat vs. raccoon/tiger}''. Using the open-vocabulary detector GroundingDINO~\cite{liu2023grounding}, we obtained 245 ID and 265 OOD predictions. For the ``refined'' prompt, we simply leveraged ChatGPT (GPT-4) to generate five nearest non-ID concepts for each ID label as exclusions, with the prompt structure and results are presented in \cref{fig:near_ood}.

\begin{table}[ht!]
  \centering
  \caption{\textbf{Near-OOD detection performance}. With more information, the ``refined'' prompt performs better on challenging cases.}
  \begin{adjustbox}{max width = \linewidth}
  \begin{tabular}{lcc}
    \toprule
        & FPR@95 ($\downarrow$) & AUROC ($\uparrow$) \\
    \midrule 
    \textbf{\ours} (\textit{Simple} prompting) & 7.79 & 96.86 \\
    \textbf{\ours} (\textit{Refined} prompting) & 5.66 & 98.03 \\ 
    
    \bottomrule
  \end{tabular}
  \end{adjustbox}
  \label{tab:near_ood}
  \vspace{-.5em}
\end{table}

\cref{tab:near_ood} presents the near-OOD performance under two prompting strategies. Further shown in \cref{fig:near_ood}, the ``refined'' prompt enables the inpainting model to generate outputs more closely aligned with the predicted labels, thereby improving near-OOD detection. These results support our hypothesis that pre-trained generative and CLIP models already possess the capability for fine-grained OOD detection, given sufficient context.

%% file: Sections/ablation.tex
\section{Ablation Studies}
\label{sec:Ablation}

\begin{figure}[htbp!]
    \centering
    \includegraphics[width=.95\linewidth]{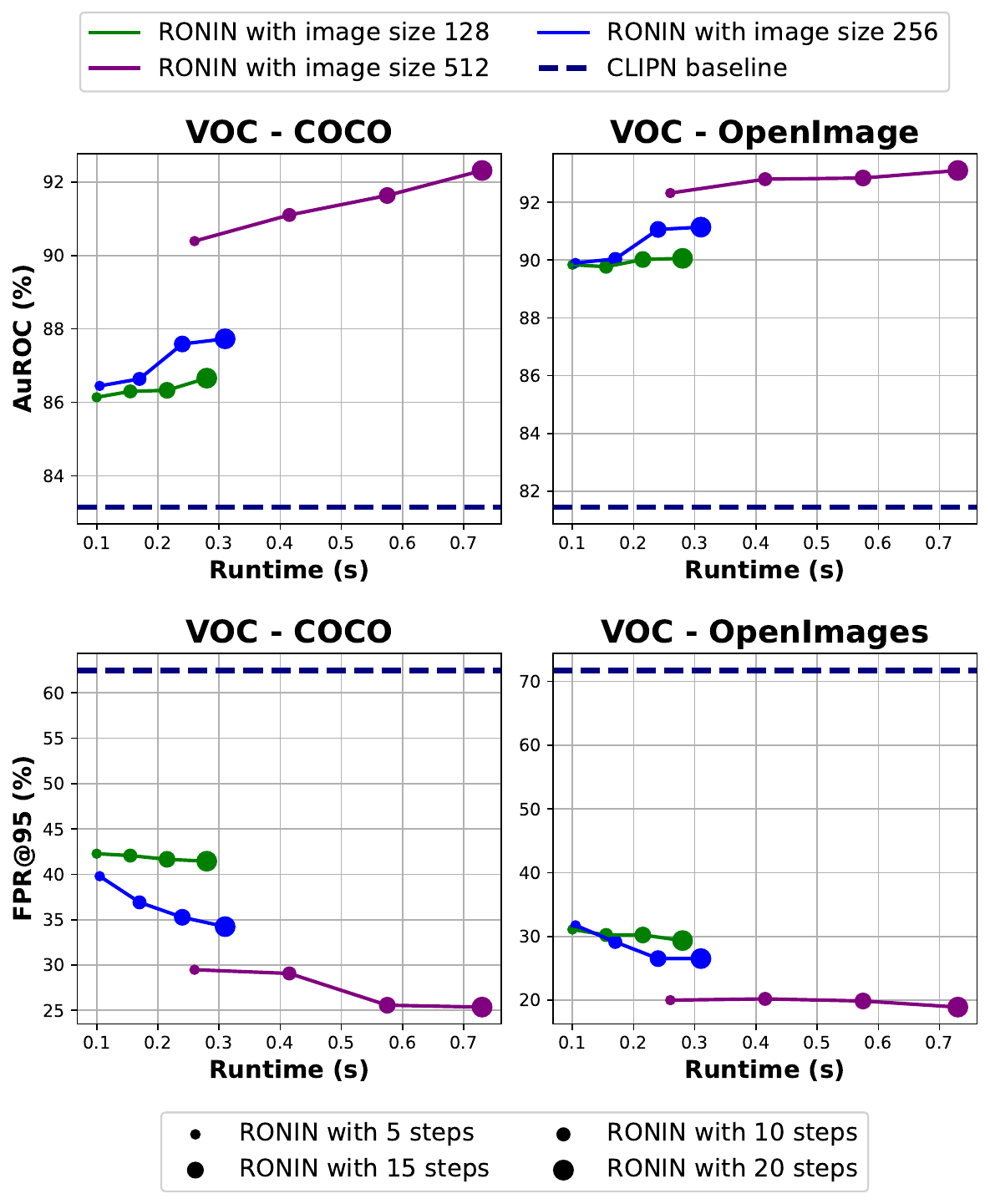}
    \caption{\textbf{Performance-speed tradeoff of \ours}. Reducing image size and denoising steps speeds up inference with minimal performance loss. At full settings, \ours achieves top accuracy with competitive runtime. CLIPN is for performance comparison.}
    \label{fig:speed}
    \vspace{-.25em}
\end{figure}

\noindent\textbf{Performance-Speed Trade off.}~~~\cref{fig:speed} analyzes the trade-off between \ours’s performance and runtime by varying the number of denoising steps and the image resolutions used during the inpainting process. For relative performance comparison, we include CLIPN~\cite{wang2023clipn} as the best-performing baseline, though we do not compare absolute runtimes between the two methods. Reducing the number of denoising steps and downscaling image sizes significantly accelerates \ours, with only minor performance degradation. These results highlight the flexibility of our approach and its potential suitability for low-resource compute environments, with reasonable performance drops.

\begin{table}[ht!]
  \centering
  \caption{\textbf{\ours with object-wise and class-wise inpainting.} Class-wise inpainting retains performance with faster runtime.}
  \begin{adjustbox}{max width = \linewidth}
      
  \begin{tabular}{lcccc}
    \toprule
    & \multicolumn{2}{c}{\textbf{FPR@95 $(\downarrow)$}} & \multicolumn{2}{c}{\textbf{AUROC $(\uparrow)$}} \\
    \cmidrule(lr){2-3}
    \cmidrule(lr){4-5}
    & Object-wise & Class-wise & Object-wise & Class-wise\\
    \midrule 
    VOC - COCO          & 26.60 & 25.36 & 91.48 & 92.31 \\
    VOC - OpenImages    & 20.00 & 18.91 & 92.68 & 93.10 \\
    BDD - COCO          & 28.89 & 26.03 & 93.74 & 92.90 \\
    BDD - OpenImages    & 26.67 & 23.33 & 92.42 & 91.90 \\
    \bottomrule
  \end{tabular}
  
  \end{adjustbox}
  \label{table:strategies_performance}
\end{table}


\noindent\textbf{Class-wise vs. Object-wise Inpainting.}~~~\cref{table:strategies_performance} compares the ``\textit{object-wise}'' inpainting strategy, which processes object individually, with our default ``\textit{class-wise}'' approach, which inpaints objects categorically. By leveraging label-wise grouping, class-wise inpainting achieves similar performances while offering more efficient computation cost. 

\begin{table}[htbp]
  \centering
  \caption{\textbf{Ablation on triplet similarity}. All three similarities are essential for effective score (\underline{underlined}). Emphasizing visual-language semantic alignment ($\alpha$) improves performance (\textbf{bolded}).}
  \begin{adjustbox}{max width = \linewidth}
      
  \begin{tabular}{lcc}
    \toprule
        & \multicolumn{2}{c}{\textbf{COCO / OpenImages}}\\
        \cmidrule(lr){2-3}
        & FPR@95 ($\downarrow$) & AUROC ($\uparrow$) \\
    \midrule 
    without $similarity(inp, \hat{y})$  & 44.54 / 37.39 & 89.77 / 91.42 \\
    without $similarity(ori, \hat{y})$  & 65.15 / 68.91 & 69.72 / 78.53 \\
    without $similarity(ori, inp)$    & \underline{30.93} / 36.96 & 86.14 / 90.10 \\
    triplet similarity, $\bm\alpha = \bm\beta = 1$  & 35.46 / \underline{24.78} & \underline{87.75} / \underline{91.43} \\
    \midrule
    triplet similarity, $\bm\alpha = 1$;  $\bm\beta = 2$  & 43.93 / 33.91 & 85.40 / 89.50 \\
    triplet similarity, $\bm\alpha = 2$; $\bm\beta = 1$  & \textbf{25.36} / \textbf{18.91} & \textbf{92.31} / \textbf{93.10} \\
    triplet similarity, $\bm\alpha = 2$; $\bm\beta = 2$   & 35.46 / 25.43 & 90.32 / 92.32\\ 

    \bottomrule
  \end{tabular}
  \vspace{-2em}
  \end{adjustbox}
  
  \label{tab:exponential_feature}
\end{table}

\noindent\textbf{Triplet Similarity Measurement.}~~~\cref{tab:exponential_feature} presents ablation studies on the triplet similarity score (\cref{equa:triplet_score}). The upper half (rows 1–4) assesses the effect of removing individual similarity components, confirming that all are essential for optimal performance. The lower half (rows 5–7) explores the weighting between semantic alignment $\textit{similarity}{(ori,\ \hat{y})}$ (weighted by $\bm{\alpha}$) and visual alignment $\textit{similarity}{(ori,\ inp)}$ (weighted by $\bm{\beta}$). As shown in \cref{fig:similarity_intuition}, emphasizing semantic alignment ($\bm{\alpha} > \bm{\beta}$) yields the best results, guiding our default choice ($\bm{\alpha} = 2$, $\bm{\beta} = 1$). However, the sensitivities of these weights suggest potential adaptability to more challenging settings.

\begin{table}[ht!]
  \centering
  \caption{\textbf{Trade-off performance with different mask sizes}. Sufficient mask size (80\% bounding box) retains cues for distinguishing object synthesis toward effective OOD performances.}
  \begin{tabular}{ccc}
    \toprule
    \multirow{2}{*}{\shortstack{Covered masking\\ratio $m$}} & \multicolumn{2}{c}{\textbf{MS-COCO}/\textbf{OpenImages}} \\
    \cmidrule(lr){2-3} 
      & FPR@95 $(\downarrow)$ & AUROC $(\uparrow)$ \\
    \midrule 
    75\%  & 30.10 / 21.74 & 90.43 / 92.41 \\
    80\%  & 25.80 / 17.74 & 91.32 / 93.84 \\
    100\% & 29.68 / 20.29 & 91.61 / 92.61 \\
    \bottomrule
  \end{tabular}
\label{table:mask_types}
\vspace{-.5em}
\end{table}

\noindent\textbf{Impact of Masking Ratios.}~~~\cref{table:mask_types} examines the impact of mask size on OOD performance. While smaller masks lead to minimal inpainting changes for effective OOD object synthesis, larger masks do not retain sufficient information for accurate ID object reconstruction, leading to suboptimal performance. More analysis is provided in \cref{app:ratio}.

\begin{table}[htbp]
  \centering
  \caption{\textbf{\ours performances with few-step generative models.} The on-par performances demonstrate \ours effectiveness with minimal reliance on specific models or synthesis quality.}
  \begin{adjustbox}{max width = \linewidth}
      
  \begin{tabular}{lccc}
    \toprule
        & \multicolumn{2}{c}{\textbf{COCO / OpenImages}}\\
        \cmidrule(lr){2-3}
        & FPR@95 ($\downarrow$) & AUROC ($\uparrow$) \\
    \midrule
    Stable Diffusion 2 (5 steps) &  27.94 / 20.00 & 92.35 / 92.32\\
    InstaFlow (1 steps) & 31.34 / 25.74 & 88.03 / 90.35\\

    \bottomrule
  \end{tabular}
  \end{adjustbox}
  
  \label{tab:different_diffusion}
  \vspace{-0.5em}
\end{table}

\noindent\textbf{Robust across Generative Models.}~~~\cref{tab:different_diffusion} evaluates \ours across various generative models, focusing on the few-step regime. \ours shows strong performance even with the one-step generative model InstaFlow \cite{liu2023instaflow}, suggesting minimal reliance on specific models or inpainting quality for effective OOD detection. Additional ablations in the many-step regime are provided in \cref{app:reliable}.

\begin{table}[ht!]
  \centering
  \caption{\textbf{Recall and falsy reject ratio over ID data}, showing \ours minimal impact on ID objects.}
  \begin{tabular}{lccc}
    \toprule
     ID set & ID object & Recall & ID Rejected \\
    \midrule 
    Pascal-VOC & 1007 & 91.57\% & 8.43\% \\
    BDD & 2632 & 96.67\% & 4.33\% \\ 
    
    \bottomrule
  \end{tabular}
  \label{tab:ID_reject}
  \vspace{-0.5em}
\end{table}

\noindent\textbf{End-to-end impact on ID object.}~~~\cref{tab:ID_reject} reports Recall and ID rejection ratios after the detection stage, showing that \ours maintains stable OOD detection without degrading ID accuracy. While BDD mainly contains small, similar vehicle categories, Pascal VOC spans far more diverse objects and animals, making it a more challenging dataset.

%% file: Sections/conclusion.tex
\section{Conclusion}
\label{sec:conclusion}

Overall, \ours delivers consistently strong performance across diverse settings, from indoor scenes to in-the-wild scenarios, demonstrating clear flexibility and generality. By exploiting subtle misalignments between generative and discriminative outputs, \ours achieves robust OOD performance without reliance on specific diffusion models or inpainting results, and it remains adaptable to a wide range of object detectors as a training-free, plug-and-play solution. Although current diffusion models limit real-time deployment, many practical applications operate in offline or post-processing pipelines where accuracy and generalizability matter most—areas where \ours excels. With strong generality and high adaptability across experiments, \ours offers a meaningful step forward for zero-shot out-of-distribution object detection.


%% file: Sections/appendix.tex
\section{Data processing}
\label{app:data}
Following \cite{du2022vos, du2022siren}, we use PascalVOC \cite{everingham2010pascal} with 20 categories\footnote{\textbf{PascalVOC ID labels}: Person, Car, Bicycle, Boat, Bus, Motorbike, Train, Aeroplane, Chair, Bottle, Dining Table, Potted Plant, TV Monitor, Couch, Bird, Cat, Cow, Dog, Horse, Sheep.} and BDD100k \cite{yu2020bdd100k} with 10 categories\footnote{\textbf{BDD100k ID labels}: Pedestrian, Rider, Car, Truck, Bus, Train, Motorcycle, Bicycle, Traffic Light, Traffic Sign.}. For OOD data, we sample from MS-COCO \cite{lin2014microsoft} and OpenImages \cite{kuznetsova2020open}, ensuring no overlap by removing all images containing ID categories. Additionally, to improve alignment with CLIP’s visual-textual representations, we standardize certain British labels (e.g., “Aeroplane”, “Couch”) to their US counterparts (“Airplane”, “Sofa”).

As our method is zero-shot and requires no training, all experiments are performed on subsets drawn from the test sets of VOC, BDD100k, COCO, and OpenImages. To ensure fairness, thoroughness, and a balanced ID–OOD distribution, we sample 200 images from the BDD100k test set and 400 images each from PascalVOC, MS-COCO, and OpenImages. The smaller sample size for BDD100k accounts for its higher object density per image.

\section{Implementation Details for Baselines}
\label{app:baseline_detail}
For discriminative zero-shot baselines, we adapt MCM~\cite{ming2022delving}, CLIPN~\cite{wang2023clipn}, TAG~\cite{liu2024tag}, OLE~\cite{ding2024zero}, and GL-MCM~\cite{miyai2025gl}—originally designed for image-level OOD detection—to the object level by cropping each detection and treating it as an individual image. Each baseline is then applied directly on top of the object detector’s predictions using its default setting, configuration, or conditioning prompt, without any specific modification. ODIN~\cite{liang2018odin} and Energy Score~\cite{liu2020energy}, developed for non-zero-shot classifier-based settings, are adapted similarly by using CLIP for zero-shot classification of detection crops. For generative baselines, we synthesize objects using Stable Diffusion 2, extract features via SimCLRv2~\cite{chen2020simclrv2}, and apply standard OOD detectors such as Mahalanobis~\cite{xiao2021we} and KNN~\cite{Sun2022knn} for fair comparison. For SIREN~\cite{du2022siren} and VOS~\cite{du2022vos}, we follow their original protocols using features from their trained detectors.

\section{Additional Ablation Studies}

\begin{figure*}[ht!]
\includegraphics[width =\textwidth]{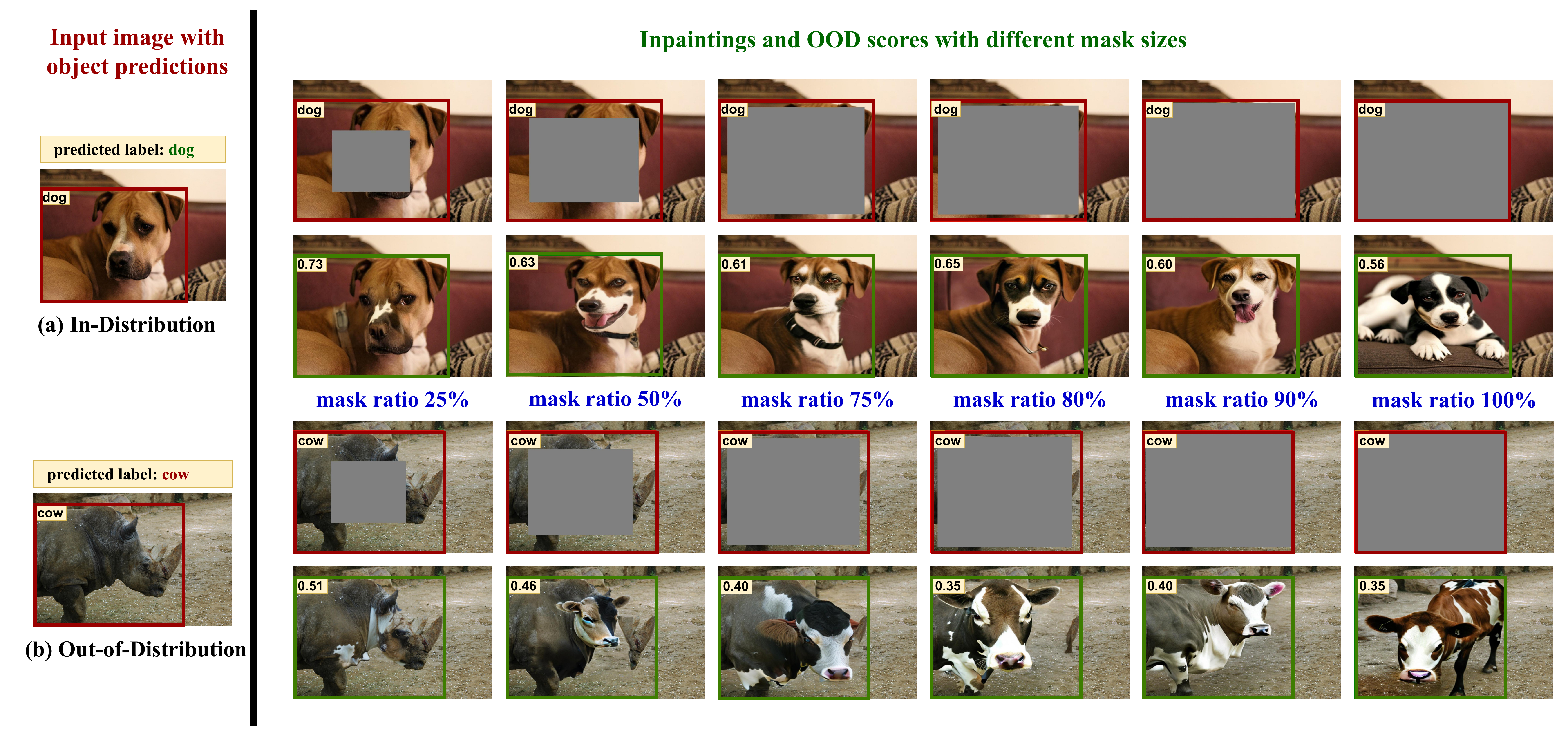}
\caption{\textbf{\ours inpainting performance across masking ratios}. Small masks preserve object structure, yielding high similarity for both ID and OOD. Large masks cause excessive distortion, lowering similarity in both cases. A balanced 80–90\% ratio retains ID integrity while amplifying OOD deviation, enabling clear ID-OOD separation.}
\label{fig:ablation_masking_visual}
\end{figure*}

\subsection{Impact of Masking Ratios}
\label{app:ratio}
Extending the ablation study in Tab. \textcolor{wacvblue}{5}, we study the impact of the masking ratio on OOD detection by varying the inpainting mask size, covering from 25\% to 100\% of the predicted bounding box. Quantitative and qualitative results in \cref{table:mask_types_appendix} and \cref{fig:ablation_masking_visual} suggest that despite being robust across masking ratios, \ours achieves optimal performance when 80–90\% of the object is masked for inpainting. As shown in \cref{fig:ablation_masking_visual}, low masking ratios preserve strong original objects' appearances in both ID and OOD cases, leading to high similarity scores. In contrast, too large a masking ratio leads to excessive deviation and ambiguity in both cases, leading to poor similarity scores. Covering 75–90\% strikes a balance, as the inpainting outcomes not only retain similar inpainting for ID samples but also produce vivid dissimilar inpaintings for OOD ones, enabling \ours to distinguish ID and OOD objects effectively.

\begin{table}[ht!]
  \centering
  \caption{\textbf{Different OOD detection performances when varying mask size for inapinting}. \ours maintains consistent and robust performance across different masking ratios, with the best performance attained when 80\%-90\% of each object is masked.}
  \begin{tabular}{ccccc}
    \toprule
    \multirow{2}{*}{\shortstack{Covered masking\\ratio $m$}} & \multicolumn{2}{c}{\textbf{MS-COCO}} & \multicolumn{2}{c}{\textbf{OpenImages}} \\
    \cmidrule(lr){2-3}  \cmidrule(lr){4-5}
      & FPR@95 $(\downarrow)$ & AUROC $(\uparrow)$ & FPR@95 $(\downarrow)$ & AUROC $(\uparrow)$  \\
    \midrule 
    25\%  & 42.47 & 88.53 & 36.74 & 90.40 \\
    50\%  & 34.23 & 89.39 & 23.91 & 92.45 \\
    75\%  & 30.10 & 90.43 & 21.74 & 92.41 \\
    80\%  & 25.80 & 91.32 & 17.74 & 93.84 \\
    90\%  & 25.57 & 91.53 & 19.57 & 92.46 \\
    100\% & 29.68 & 91.61 & 20.29 & 92.61 \\
    \bottomrule
  \end{tabular}
\label{table:mask_types_appendix}
\end{table}

\begin{table*}[htbp]
  \centering
  \caption{\textbf{\ours performances across different diffusion models.} The consistently robust performances across inpainting models, including one-step inpainting, demonstrate \ours effectiveness without any dependence on specific models or inpainting quality.}
  \begin{adjustbox}{max width = \linewidth}
      
  \begin{tabular}{llcccc}
    \toprule
        &&\multirow{2}{*}{\shortstack{\textbf{Denoising Process}\\(steps)}} & \multicolumn{2}{c}{\textbf{COCO / OpenImages}}& \multirow{2}{*}{\shortstack{\textbf{Avg runtime/image}\\(seconds)}}  \\
        \cmidrule(lr){4-5}
        && & FPR@95 ($\downarrow$) & AUROC ($\uparrow$) & \\
    \midrule 
    \multirow{3}{*}{\textbf{Many-step Regime}}&Kandinsky 2.1 \footnote{\url{https://huggingface.co/kandinsky-community/kandinsky-2-1-inpaint}} & 20 & 28.99 / 23.96 & 89.92 / 91.62 & 1.12 \\
    &Stable Diffusion 1.5 \footnote{\url{https://huggingface.co/stable-diffusion-v1-5/stable-diffusion-inpainting}}  & 20 & 29.90 / 23.70  & 91.36 / 92.36 & 1.20\\
    &Stable Diffusion 2 (\textbf{\textit{default choice}})   & 20 & {25.84} / {18.91} & {92.28} / {93.30} & 0.83  \\
    &Stable Diffusion 3 \footnote{\url{https://huggingface.co/IrohXu/stable-diffusion-3-inpainting}}  & 20 & 21.09 / 18.89 & {94.01} / {94.43} & 2.86  \\
    \midrule
    \multirow{2}{*}{\textbf{Few-step Regime}}&Stable Diffusion 2 & 5 & 27.94 / 20.00 & 92.35 / 92.32 & 0.19\\
    &InstaFlow & 1 & 31.34 / 25.74 & 88.03 / 90.35 & 0.03\\

    \bottomrule
  \end{tabular}
  \end{adjustbox}
  
  \label{tab:different_diffusion_appendix}
\end{table*}

\subsection{Robust across Generative Models}
\label{app:reliable}
Extending the ablation study in Tab. \textcolor{wacvblue}{6} to the many-step regime, we further evaluate \ours by replacing the default Stable Diffusion 2 with Kandinsky 2.1 \cite{razzhigaev2023kandinsky} and Diffusion Model 1, both using 20 denoising steps. We also include results from the few-step regime study, with the one-step diffusion model InstaFlow~\cite{liu2023instaflow} and 5-step Stable Diffusion 2 for comparison. As shown in \cref{tab:different_diffusion_appendix}, \ours remains consistently robust across different diffusion models and configurations. Notably, Stable Diffusion 2 provides the best trade-off between performance and runtime, even in the few-step generative mode. These findings suggest that by subtly leveraging semantic misalignment between original and synthesized objects for zero-shot OOD detection, \ours performs stably and reliably without relying on any specific diffusion model or high-quality inpainted outputs. Even when models produce diverse inpainting results, our framework remains both robust and effective.

\clearpage